\documentclass[11pt]{article}

\usepackage[margin=1.1in]{geometry}
\usepackage{newtxtext}
\usepackage{amsmath,amssymb,amsthm}
\usepackage{newtxmath}
\usepackage{graphicx}
\usepackage{booktabs}
\usepackage{array}
\usepackage{microtype}
\usepackage{needspace}
\usepackage[hidelinks]{hyperref}

\newtheorem{proposition}{Proposition}
\newtheorem{corollary}{Corollary}

\newtheorem{assumption}{Assumption}
\newtheorem{remark}{Remark}

\newcommand{\dstar}{d^{*}}
\newcommand{\ur}{U_{\mathrm{R}}}
\newcommand{\up}{U_{\mathrm{P}}}
\newcommand{\ua}{U_{\mathrm{A}}}
\newcommand{\cR}{c_{\mathrm{R}}}
\newcommand{\cP}{c_{\mathrm{P}}}
\newcommand{\dR}{\delta_{\mathrm{R}}}
\newcommand{\dP}{\delta_{\mathrm{P}}}
\newcommand{\cdiag}{c_{\mathrm{diag}}}

\title{When AI Agents Should Repair, Replan, or Abandon:\\[2pt]\large A Threshold Theory of Plan Commitment}
\author{Gaurav Tewari\\\small Omega Venture Partners\\\small \href{https://www.omegavp.com/}{www.omegavp.com}}
\date{November 2026}

\begin{document}

\maketitle

\begin{abstract}
\noindent
AI agents often discover mid-execution that their plan is broken, for example when an action fails or a tool is withdrawn. Deployed systems use conventions such as retry loops or ``always replan,'' and a recent benchmark finds no fixed strategy dominates. This paper derives the optimal policy. In the model, a disruption destroys the causal support of part of a plan; repair is a partial reset that preserves the plan's structure, replanning is a full reset, and changing a plan is costly because others have conditioned on it. The optimal policy is a threshold in effective disruption depth: repair shallow damage, replan after deep damage, and abandon the goal when even a fresh plan is not worth its cost. Depth depends on plan structure. In expectation, a single broken action forces $1+R(G)/n$ actions to be revalidated, where $R(G)$ counts the pairs of actions joined by a dependency path; to first order, a fully serial plan switches to replanning at raw damage about $w$ times smaller than a plan of $w$ parallel chains. A closed-form stability-price threshold determines when a worse repaired plan should be kept over a better fresh one. As language-model planners push replanning costs toward zero, the stability price becomes the only force sustaining plan commitment. Under repeated disruptions, revision frictions make commitment hysteresis optimal: the agent keeps executing a plan it would never choose fresh, and each repair adds patch debt until the debt outweighs repair's cost advantage and the agent rebuilds. A value-of-diagnosis result states when measuring damage is worth its compute and shows that ``always replan'' is optimal when damage is predictably deep and replanning is the better fallback. A separately implemented numerical checker tests each proposition (zero violations; $1.6$ million static-model evaluations), and the model's regime map reproduces that benchmark's regime flips qualitatively.
\end{abstract}

\vspace{0.5em}
\noindent\textbf{Keywords:} planning, plan repair, replanning, agent architectures, commitment, hysteresis, value of information, decision theory, AI agents.

\section{Introduction}\label{sec:intro}

A deployed AI agent will at some point find that the plan it is executing no longer matches the world. A booked flight is canceled, or an API the agent meant to call starts refusing its requests. The agent then has three options: \emph{repair} the current plan and keep what still works, \emph{replan} from scratch, or \emph{abandon} the goal and escalate. Current systems choose among them by convention. Some scaffolds replan after every disturbance or retry a fixed number of times, and hand-tuned hysteresis guards have been proposed to stop agents from thrashing between alternatives \cite{irrational2026}. Recent evidence indicates that no fixed strategy is right in general. A 2026 benchmark of travel-planning agents under resource disruptions finds that full replanning, hierarchical repair, and local editing each win in some regimes and lose in others, and its authors close with practical guidelines \cite{yuan2026replan}.

The question has a long history in AI planning. Nebel and Koehler treated it as a computational problem and proved that reusing an existing plan is, in the worst case, no cheaper than planning from scratch, and that conservative modification can be strictly harder \cite{nebel1995}. Fox, Gerevini, Long, and Serina then studied it empirically \cite{fox2006}. They introduced \emph{plan stability}, the distance between an old plan and its replacement; argued informally that stability is valuable because observers and cooperating agents rely on predictable plans; and showed that a repair strategy built on the LPG planner \cite{gerevini2003} preserves far more of the original plan than replanning does. Other work refined repair algorithms and their metrics \cite{vanderkrogt2005,saetti2025,babli2023}, defined variants of the replanning problem without optimality results \cite{talamadupula2014}, and compared repair strategies empirically in hierarchical planning \cite{zaidins2025}. None of this work says when repair, replanning, or abandonment is optimal as a function of measurable properties of the plan, the damage, and the deployment context; Yuan et~al.\ limit their guidelines to the settings they evaluated \cite{yuan2026replan}.

This paper develops a decision theory for that choice from three modeling assumptions. First, damage propagates through the plan. A disruption removes the causal support of every action that depended on the actions it hits, so the quantity that matters for the decision is the \emph{effective depth} of the damage: raw damage amplified through the plan's dependency graph. Second, repair and replanning are different kinds of instrument. Repair is a partial reset whose cost scales with the damage and whose output stays close to the incumbent plan. Replanning resets the plan fully; its cost scales with the size of the problem, and its output typically lies far from the incumbent in plan distance. Third, plan distance is costly in itself. When humans supervise an agent, other agents coordinate with it, or downstream commitments rest on its announced intentions, each action changed by a revision imposes a coordination cost. This is the stability argument of Fox et~al.\ \cite{fox2006}, given a price $\sigma$ per unit of plan distance and placed inside the agent's objective.

Our first result, Proposition~\ref{prop:threshold}, gives the basic policy. Under mild monotonicity assumptions, the optimal choice is a threshold $\dstar$ in effective depth: repair below it, and replan or abandon above it. The proposition classifies the possible regimes, including a \emph{stability lock-in} regime in which repair dominates replanning at every depth. Corollary~\ref{cor:scale} shows that, under a stated condition, the threshold rises with problem scale. This accounts for a regularity reported by Fox et~al.\ \cite{fox2006}: repair's advantage grows with problem size, as expected when repair cost scales with the damaged region and replanning cost scales with the whole problem.

Plan structure enters through Proposition~\ref{prop:topology}. For a broken action chosen uniformly at random, the expected validation footprint is $1+R(G)/n$ actions, where $R(G)$ is the number of ordered pairs of actions joined by a dependency path. This amplification factor ranges from $1$ for independent actions to $(n{+}1)/2$ for a fully serial plan. Plan topology therefore moves the threshold alongside the cost parameters, and parallelizing a plan is a design choice whose effect on commitment can be quantified.

Proposition~\ref{prop:sigma} gives, in closed form, the stability price $\sigma^{*}$ above which a strictly worse repaired plan should still be kept in preference to replanning, and Corollary~\ref{cor:cheap} takes the limit in which replanning becomes free, a limit that matters more as language-model planners drive replanning costs down. Cushing and Kambhampati argued informally that the usual reasons for preserving a plan's structure reduce to the presence of other agents \cite{cushing2005}; to our knowledge, the corollary is the first formal statement that \emph{when replanning is free, the only reason to keep a plan is that others depend on it}.

Under repeated disruptions, revision frictions give the optimal policy an inaction band (Proposition~\ref{prop:hysteresis}), which we call \emph{commitment hysteresis} in the sense the term has in the economics of fixed adjustment costs \cite{dixit1992,stokey2009}. An optimally managed agent keeps executing a degraded plan that it would never choose fresh and, under a stated condition, revises once mismatch crosses a threshold. Each repair adds to a ``patch debt'' that constrains later adaptation; once the cost of that debt outweighs repair's cost advantage, rebuilding replaces repair at the next revision.

Diagnosis is itself a costly choice (Proposition~\ref{prop:voi}). Measuring the damage before deciding has positive value only when the prior over depth puts mass on both sides of the threshold. When damage is predictably deep and replanning is the better fallback, diagnosis has no value and ``always replan'' is optimal, so the model recovers that convention as a special case.

Each proposition is tested by a numerical checker that recomputes from the model's primitives and does not reuse the paper's algebra (Section~\ref{sec:numerics}). The tests comprise a symbolic audit of the closed forms; randomized scans of $4{,}000$ drawn model instances, with zero violations of the claimed monotonicity, uniqueness, or regime structure; exact graph enumeration of the topology results on $600$ random DAGs; dynamic-programming solutions of the sequential model for $59$ parameter draws, all with threshold structure; and $3{,}000$ value-of-information instances.

The theory concerns when repair is worth doing and says nothing about how to repair, which is the subject of \cite{gerevini2003,saetti2025,likhachev2005}. The stability price $\sigma$ is a reduced form for coordination, oversight, and commitment costs, and we do not model the strategic interaction that generates it. The analysis therefore stays within single-agent planning and decision-making under uncertainty.

\section{Related work}\label{sec:related}

\paragraph{Plan repair and replanning in AI.}
The computational comparison begins with Nebel and Koehler \cite{nebel1995} and the empirical comparison with Fox et~al.\ \cite{fox2006}, who measure plan stability as action-set distance and argue informally that it matters for predictability and graceful execution; in their DriverLog-Time runs, repair changed $8.9\%$ of the original plan's first-step actions and replanning $66.0\%$. Other work made repair optimal with respect to stability \cite{saetti2025}, extended repair to multi-agent and hierarchical task network (HTN) settings \cite{komenda2012,zaidins2025}, and framed replanning in terms of sensitivity to commitments \cite{cushing2005,vanderkrogt2005}. Talamadupula et~al.\ distinguish restart, similarity-constrained, and commitment-sensitive replanning \cite{talamadupula2014}. Babli et~al.\ use the phrase ``plan commitment'' for a repair heuristic and identify the choice of revision thresholds as an open question \cite{babli2023}. Incremental heuristic search provides fast repair algorithms for robotics without a break-even theory \cite{stentz1995,koenig2002,likhachev2005}, and dynamic constraint satisfaction reuses solutions in a similar way \cite{verfaillie1994}. The closest empirical work is the 2026 benchmark of Yuan et~al.\ \cite{yuan2026replan}, which compares three revision strategies under systematic disruptions, finds that no fixed strategy dominates, and leaves a minimal-edit objective unimplemented.

\paragraph{Intention reconsideration.}
The belief--desire--intention (BDI) tradition asks when an agent should \emph{reconsider} its intentions at all. Work in this line compares bold and cautious agents in dynamic environments \cite{kinny1991}, treats reconsideration as meta-level control of deliberation in the framework of \cite{russell1991}, and derives decision-theoretic reconsideration policies \cite{schut2001,schut2004}. This literature studies whether to deliberate again. It does not model a choice among structurally different revision instruments, the salvage value of the current plan, or a price on stability. Horty and Pollack \cite{horty2001} evaluate new options in the context of existing plans; they do not address the revision of a broken plan. Bratman's account of intentions as commitment devices \cite{bratman1987} is the philosophical root of our stability price.

\paragraph{Repair-or-replace theory in operations research.}\label{par:or}
Operations research has several threshold theorems that resemble Proposition~\ref{prop:threshold}. Repair-cost-limit policies repair a failed machine when its estimated repair cost is below a limit and replace it otherwise \cite{hastings1969,dohi2001}; repair-\emph{number}-limit policies replace it at the $k$th minimal repair \cite{park1979,nakagawa1981}; and virtual-age models describe imperfect repair \cite{kijima1989}. These models act on a scalar machine state subject to physical wear. They have no analogue of a plan's internal dependency structure, they attach no cost to the replacement being \emph{different} (a replaced machine inconveniences no one by changing), and they have a single repair instrument. The patch-debt dynamics of Section~\ref{sec:dynamic} are closest to the repair-number-limit policies and differ from them in these three respects. Match-up scheduling proves conditions under which it is optimal, after a disruption, to reschedule so as to rejoin the preschedule at a future horizon \cite{bean1991}. There the state is a schedule with release dates and convex tardiness costs, the decision is when to rejoin, and nothing corresponds to dependency amplification, priced stability, or a choice between two structurally different revision instruments. Rescheduling frameworks classify strategies without analyzing optimality \cite{vieira2003}. The results here that have no counterpart in this literature are the ones specific to plans: the amplification law (Proposition~\ref{prop:topology}), the stability price and its cheap-planning limit (Proposition~\ref{prop:sigma}), the two-instrument patch-debt dynamics (Proposition~\ref{prop:hysteresis}), and the diagnosis layer (Proposition~\ref{prop:voi}).

\paragraph{Agent plan revision.}
Recent work on autonomous agents, including agents built on large language models (LLMs), addresses the behavior analyzed here with hysteresis guards against plan thrashing \cite{irrational2026}, budgeted replanning architectures \cite{cascade2026}, stopping rules for verify--repair loops \cite{vrr2026}, and benchmarks of recovery cost after tool failure \cite{whentoolsfail2026}. A 2026 survey lists constraint-satisfying plan repair among open problems \cite{survey2026}.

\section{The static model}\label{sec:model}

\subsection{Primitives}

A plan is a finite set of actions $A$, $|A| = n \ge 1$, partially ordered by causal dependencies: the directed acyclic graph $G = (A, E)$ has an edge $u \to v$ when $v$'s preconditions rely on $u$'s effects. Executing the plan to completion is worth $V > 0$. We normalize $V$ to be the value of the \emph{best plan available from the post-disruption state}, so that successful replanning attains $V$ before costs and every quality penalty borne by repair appears as a gap $\Delta \ge 0$ relative to the fresh optimum. This normalization deliberately biases the model \emph{against} repair, which nevertheless remains optimal over large regions of the parameter space.

A disruption makes a set $B \subseteq A$ of remaining actions inexecutable, and the damage propagates. The \emph{validation footprint} of $B$ is its forward cone
\begin{equation}
I(B) \;=\; B \,\cup\, \{\, v \in A : u \leadsto v \text{ for some } u \in B \,\},
\label{eq:cone}
\end{equation}
where $u \leadsto v$ denotes reachability in $G$. Every action whose causal support runs through a broken action must at least be revisited. The footprint is a worst case for actual invalidation. When a broken action has an effect-equivalent substitute, repair preserves its dependents and the decision-relevant damage is smaller. We write $b = |B|/n$ for \emph{raw damage} and $d = |I(B)|/n \in [0,1]$ for \emph{effective depth}. As a measure of decision-relevant damage, $d$ is exact when no effect-equivalent substitutes exist and an upper bound otherwise. Section~\ref{sec:topology} characterizes the map from $b$ to $d$; until then, $d$ is the primitive.

\needspace{15\baselineskip}
The agent chooses among three responses.
\begin{itemize}
\item \textbf{Repair}: a structure-preserving revision. It costs $\cR(d)$ (computation and delay scale with the damaged region), achieves completion value $V - \Delta(d)$ (adaptation is constrained search near the incumbent plan \cite{nebel1995,fox2006}), and produces a plan at distance $D_{\mathrm{R}} = \dR(d)\, n$ from the incumbent, where distance is the symmetric action-set difference of \cite{fox2006}.
\item \textbf{Replan}: fresh synthesis. It costs $\cP$, achieves $V$, and lands at distance $D_{\mathrm{P}} = \dP\, n$, with $\dP$ typically large (in \cite{fox2006}, replanning replaced the majority of the plan; in \cite{yuan2026replan}, replanning changed $6$--$9$ itinerary slots against $1$--$2$ for repair).
\item \textbf{Abandon}: take an outside option worth $W \ge 0$, such as escalating to a human, decomposing the goal differently, or giving up.
\end{itemize}

The last primitive is a price on plan distance. Others condition on the incumbent plan: human overseers may have approved it, and cooperating agents and downstream commitments may depend on it. Let $\sigma \ge 0$, the \emph{stability price}, be the cost a revision imposes per unit of plan distance. It is a reduced form for coordination, oversight, and commitment costs, in the spirit of \cite{fox2006,bratman1987}, and $\sigma = 0$ recovers the frictionless special case.

The payoffs are
\begin{align}
\ur(d) &= V - \Delta(d) - \cR(d) - \sigma\, \dR(d)\, n, \label{eq:UR}\\
\up    &= V - \cP - \sigma\, \dP\, n, \label{eq:UP}\\
\ua    &= W. \label{eq:UA}
\end{align}

\begin{assumption}\label{ass:a1}
$\cP > 0$; $\cR$, $\Delta$, $\dR$ are continuous and strictly increasing on $[0,1]$ with $\cR(0)=\Delta(0)=\dR(0)=0$; $\dR(d) \le \dP$ for all $d$; and $\cR(1) + \Delta(1) \ge \cP$ (repairing total destruction is weakly worse than a fresh start, since at $d=1$ repair is replanning with extra constraints).
\end{assumption}

Appendix~\ref{app:general} adds success probabilities $q_{\mathrm{R}}(d), q_{\mathrm{P}} \in (0,1]$ and a fallback after failed repair; the threshold structure carries over. Table~\ref{tab:notation} collects the notation.

\begin{table}[t]
\centering\small
\begin{tabular}{@{}lll@{}}
\toprule
Symbol & Meaning & Source in deployment \\
\midrule
$n$, $G$ & remaining plan size; dependency DAG & plan representation \\
$B$, $b$ & broken actions; raw damage $|B|/n$ & failed precondition checks \\
$I(B)$, $d$ & validation footprint \eqref{eq:cone}; effective depth $|I(B)|/n$ & reachability over $G$ \\
$\alpha(G)$ & damage amplification \eqref{eq:alpha} & computable at plan time \\
$V$, $W$ & post-disruption fresh optimum; outside option & assigned \\
$\cR(d)$, $\cP$ & repair / replanning compute cost & token and latency logs \\
$\Delta(d)$ & repair's quality gap to the fresh optimum & paired audits (estimated) \\
$\dR(d)$, $\dP$ & revision distance per action (repair / replan) & action-set metric \cite{fox2006} \\
$\sigma$ & stability price per unit plan distance & policy knob (assigned) \\
$x$, $y$ & accumulated mismatch; patch debt (Section~\ref{sec:dynamic}) & monitoring; revision log \\
$F$, $\cdiag$ & prior over depth; diagnosis cost (Section~\ref{sec:voi}) & disruption history; compute \\
\bottomrule
\end{tabular}
\caption{Model notation and the source of each quantity in a deployed agent framework: logged directly, computed from the plan, or estimated or assigned (Section~\ref{sec:implications}).}
\label{tab:notation}
\end{table}

\subsection{The commitment threshold}\label{sec:threshold}

Define the repair premium $\varphi(d) = \ur(d) - \up = \bigl[\cP - \cR(d)\bigr] - \Delta(d) + \sigma n \bigl[\dP - \dR(d)\bigr]$ and the lock-in boundary
\begin{equation}
\sigma_{\mathrm{lock}} \;=\;
\begin{cases}
\dfrac{\cR(1) + \Delta(1) - \cP}{\,n\,[\dP - \dR(1)]\,} & \text{if } \dR(1) < \dP,\\[8pt]
+\infty & \text{if } \dR(1) = \dP \text{ and } \cR(1)+\Delta(1) > \cP \quad (\text{lock-in unattainable}),\\[2pt]
0 & \text{if } \dR(1) = \dP \text{ and } \cR(1)+\Delta(1) = \cP \quad (\varphi(1)=0 \text{ at every } \sigma).
\end{cases}
\label{eq:lock}
\end{equation}

\needspace{14\baselineskip}
\begin{proposition}[Threshold policy and regimes]\label{prop:threshold}
Under Assumption~\ref{ass:a1}:
\begin{enumerate}
\item[(i)] $\varphi$ is strictly decreasing with $\varphi(0) = \cP + \sigma n \dP > 0$. If $\sigma < \sigma_{\mathrm{lock}}$ there is a unique $\dstar \in (0,1]$ with $\varphi(\dstar) = 0$: repair is optimal against replanning exactly on $d < \dstar$.
\item[(ii)] \emph{(Stability lock-in.)} If $\sigma \ge \sigma_{\mathrm{lock}}$, repair dominates replanning at every depth $d \in [0,1]$.
\item[(iii)] \emph{(Full policy.)} In the interior regime ($\sigma < \sigma_{\mathrm{lock}}$), if $\up \ge W$ the optimal policy is to repair on $[0,\dstar)$ and replan on $(\dstar, 1]$, with indifference at $\dstar$; under lock-in with $\up \ge W$, repair everywhere. If $\up < W$, replanning is never used: repair while $\ur(d) > W$ and abandon otherwise. That is, repair on $[0, d^{\circ})$ and abandon on $(d^{\circ},1]$ when $\ur(1) \le W < V$, where $d^{\circ}$ is the unique root of $\ur(d^{\circ}) = W$; repair at every depth when $\ur(1) > W$; and abandon everywhere when $W \ge V$.
\item[(iv)] \emph{(Comparative statics.)} In the interior regime, $\dstar$ is strictly increasing in $\sigma$ and $\cP$, and strictly decreasing under pointwise upward shifts of $\Delta$ or $\cR$, and of $\dR$ when $\sigma > 0$.
\end{enumerate}
\end{proposition}

\begin{figure}[t]
\centering
\includegraphics[width=0.72\linewidth]{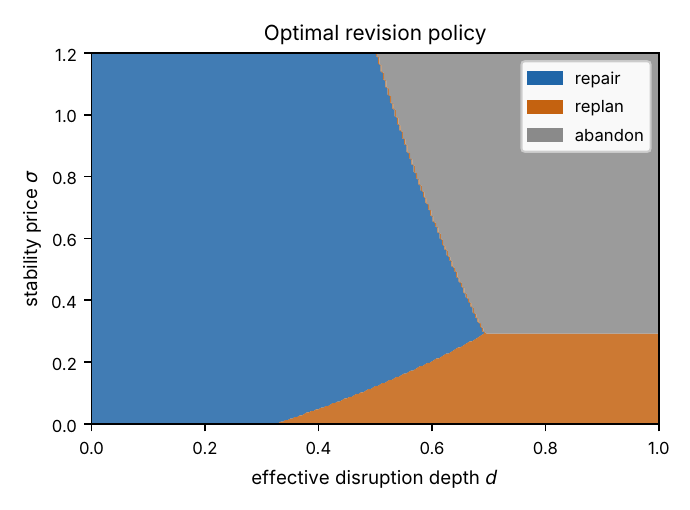}
\caption{Optimal revision policy in (effective depth $d$, stability price $\sigma$), computed from \eqref{eq:UR}--\eqref{eq:UA} at the baseline parameters of Section~\ref{sec:numerics}. The repair--replan boundary $\dstar(\sigma)$ increases in $\sigma$ (Proposition~\ref{prop:threshold}(iv)). At high $\sigma$, replanning's larger plan distance makes abandonment the relevant alternative.}
\label{fig:policy}
\end{figure}

Proofs are in Appendix~\ref{app:proofs}. The comparative statics in $\sigma$ follow because each unit increase in $\sigma$ raises $\varphi$ pointwise by $n[\dP - \dR(d)] \ge 0$, so stability subsidizes whichever instrument moves the plan less, which under Assumption~\ref{ass:a1} is always repair. Figure~\ref{fig:policy} shows the three policy regions. Abandonment is optimal where $\sigma$ is high enough that replanning is worth less than the outside option and the damage is deep enough that repair is worth less as well.

\begin{corollary}[Scale widens the repair region]\label{cor:scale}
Let repair cost scale with the damaged region and replanning cost with the problem: $\cR(d) = \kappa\, h(dn)$ and $\cP = \kappa\, C(n)$ with $h, C$ increasing, and adopt the convention $\dstar = 1$ under lock-in. For admissible sizes $n' > n$, if $\kappa [C(n') - h(dn')] + \sigma n' [\dP - \dR(d)] \;\ge\; \kappa [C(n) - h(dn)] + \sigma n [\dP - \dR(d)]$ for all $d \in (0,1)$, then $\dstar(n') \ge \dstar(n)$, strictly when $\dstar(n) < 1$ and the inequality is strict at $\dstar(n)$.
\end{corollary}

The condition says that replanning's total burden grows with scale at least as fast as repair's. It holds, for instance, for $C(n) = n$ and $h(z) = z$ with $\Delta$ fixed. The corollary turns the observation of Fox et~al.\ that repair pays most on large problems \cite{fox2006} into a prediction of the model.

\subsection{Structural damage amplification}\label{sec:topology}

Effective depth is determined by raw damage and the plan's dependency structure. Reachability-based impact sets are standard in software change impact analysis \cite{bohner1996,yau1980,law2003}. We derive their exact expected size on plan graphs, characterize the extremal graphs, and trace the consequences for commitment thresholds. For a single broken action drawn uniformly at random, the expected footprint has a closed form.

\begin{proposition}[Amplification law]\label{prop:topology}
Let $R(G)$ be the number of ordered pairs $(u,v)$, $u \neq v$, with $u \leadsto v$ in $G$. For $B = \{a\}$ with $a$ uniform on $A$,
\begin{equation}
\alpha(G) \;\equiv\; \mathbb{E}\,|I(B)| \;=\; 1 + \frac{R(G)}{n},
\label{eq:alpha}
\end{equation}
and $1 \le \alpha(G) \le \tfrac{n+1}{2}$, with the lower bound attained exactly by the edgeless plan (independent actions) and the upper bound exactly by plans whose transitive closure is a total order (fully serial plans). For $w$ parallel chains of length $n/w$, $\alpha = \tfrac{1}{2}(n/w + 1)$. For multi-action damage, $|I(B)| \le \min\{n, \sum_{a \in B} |I(\{a\})|\}$ pointwise, hence $\mathbb{E}|I(B)| \le \min\{n, \mathbb{E}\sum_{a \in B} |I(\{a\})|\}$ by Jensen's inequality ($\min\{n,\cdot\}$ is concave).
\end{proposition}

\begin{corollary}[Serial plans are brittle]\label{cor:serial}
With effective depth approximated as $d \approx \alpha(G)\, b$ for small raw damage $b$, a fully serial plan reaches any given depth threshold at raw damage approximately $w$ times smaller than a plan of the same size made of $w$ parallel chains. Parallelizing a plan therefore protects commitment: reducing $\alpha$ raises, in proportion, the raw damage that falls inside the repair region, up to the footprint bound.
\end{corollary}

\begin{figure}[t]
\centering
\includegraphics[width=0.72\linewidth]{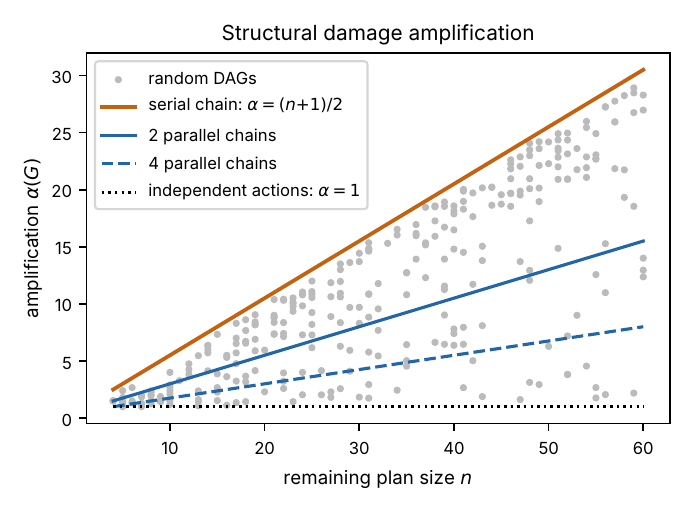}
\caption{Amplification $\alpha(G)$ by plan structure: exact values for serial chains ($\alpha = (n{+}1)/2$), two and four parallel chains, independent actions ($\alpha = 1$), and $300$ random DAGs (gray; a separate set of $600$ random DAGs was checked against \eqref{eq:alpha}), all computed by direct reachability enumeration. No DAG exceeds the serial bound, as Proposition~\ref{prop:topology} requires.}
\label{fig:topology}
\end{figure}

In Proposition~\ref{prop:topology}, the state that drives the decision is a graph, and the shape of the graph changes the optimal policy. (Change impact analysis uses the same footprint construct \cite{bohner1996} to measure maintenance effort, and in design-structure analysis $\alpha(G)/n$ is the ``propagation cost'' of a system \cite{maccormack2006}. To our knowledge, the extremal characterization and the link to revision thresholds are new.) The proposition also yields a testable prediction for benchmarks such as that of \cite{yuan2026replan}, whose itineraries contain serial temporal dependencies (a replacement flight may shift the arrival time and prevent a scheduled visit). In serial domains, replanning should become competitive at modest raw damage; in agentic workloads with parallel subtasks, repair should remain optimal up to proportionally larger raw damage.

\subsection{The price of predictability}\label{sec:sigma}

We now ask when stability justifies keeping a strictly worse plan. Suppose the quality-and-cost calculus alone favors replanning, so that $\Delta(d) > \cP - \cR(d)$.

\begin{proposition}[Exact stability threshold]\label{prop:sigma}
For any $d$ with $\Delta(d) > \cP - \cR(d)$ and $\dR(d) < \dP$, repair beats replanning ($\ur(d) \ge \up$) if and only if
\begin{equation}
\sigma \;\ge\; \sigma^{*}(d) \;=\; \frac{\Delta(d) + \cR(d) - \cP}{\,n\,[\dP - \dR(d)]\,}.
\label{eq:sigmastar}
\end{equation}
\end{proposition}

\begin{corollary}[Cheap planning erodes all commitment except coordination]\label{cor:cheap}
As $\cP \to 0$, $\sigma^{*}(d) \to [\Delta(d) + \cR(d)] / (n[\dP - \dR(d)]) > 0$ for every $d \in (0,1]$ with $\dR(d) < \dP$. If in addition $\sigma = 0$, then $\dstar \to 0$: when a fresh plan beats the outside option ($\up \ge W$), the optimal frictionless agent replans at \emph{every} disturbance.
\end{corollary}

\begin{remark}[Lumpy re-approval]\label{rem:tariff}
Real coordination costs are often two-part. Any revision triggers a fixed re-review cost $\sigma_{0}$, and each changed action adds a marginal cost $\sigma_{1}$, so that $\sigma(D) = \sigma_{0}\mathbf{1}\{D>0\} + \sigma_{1} D$. In the static model both instruments revise, so $\sigma_{0}$ enters $\ur$ and $\up$ equally and cancels from $\varphi$. The repair--replan threshold of Proposition~\ref{prop:threshold} is therefore unchanged, while comparisons with abandonment shift, because $W$ carries no re-review cost. The results that price the \emph{differential} plan distance, namely $\sigma^{*}$, $\sigma_{\mathrm{lock}}$, and Corollary~\ref{cor:cheap}, depend on the marginal component and require $\sigma_{1} > 0$. In the sequential model of Section~\ref{sec:dynamic}, continuing avoids $\sigma_{0}$ entirely, so a lumpy tariff acts as a higher revision friction $\kappa$, which widens the band in the two-period model (Appendix~\ref{app:twoperiod}) and did not narrow it in any of the $25$ draws of Section~\ref{sec:dynamic}.
\end{remark}

\begin{remark}[Weakly monotone primitives]\label{rem:weak}
For LLM-based agents, $\cR$ may be nearly flat (one repair call uses a roughly constant number of tokens), and $\Delta$ may vanish over a range of shallow damage. Relaxing Assumption~\ref{ass:a1} to nondecreasing primitives turns the threshold $\dstar$ into an indifference interval and changes the churn limit of Corollary~\ref{cor:cheap} to $\dstar \to \inf\{d : \cR(d) + \Delta(d) > 0\}$, which can be strictly positive: with free shallow repair, even a frictionless agent patches small damage. The qualitative classification of regimes is unchanged; strictness and uniqueness are lost.
\end{remark}

\begin{figure}[t]
\centering
\includegraphics[width=0.72\linewidth]{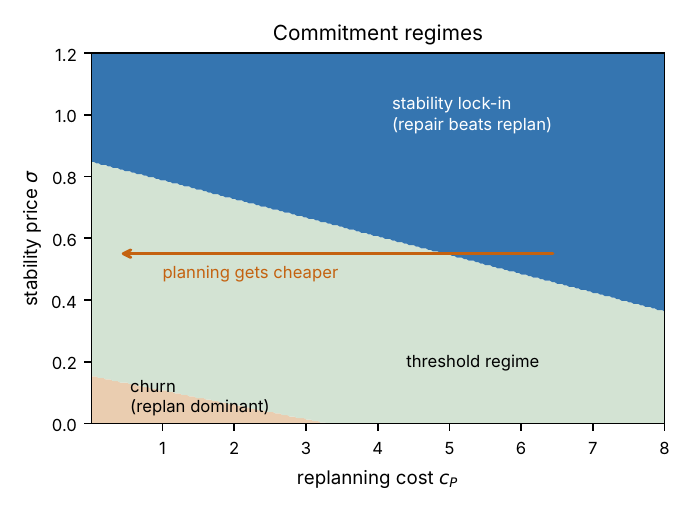}
\caption{Commitment regimes in (replanning cost $\cP$, stability price $\sigma$) at the baseline parameters of Section~\ref{sec:numerics}: churn ($\dstar < 0.35$, so damage at most depths triggers replanning), the interior threshold regime, and stability lock-in ($\sigma \ge \sigma_{\mathrm{lock}}$; repair dominates replanning at every depth). At low $\cP$, the regime depends on $\sigma$, which the architecture sets.}
\label{fig:regimes}
\end{figure}

By Corollary~\ref{cor:cheap}, when replanning is free, the priced coordination cost of plan churn is the only force sustaining plan commitment. As language-model planners reduce the marginal cost of a fresh plan, the model predicts that a quality-maximizing agent will revise its plan at nearly every disturbance unless the architecture puts a price on plan churn. Under cheap planning, an agent framework therefore has to measure plan stability, for which the distance metric of \cite{fox2006} suffices, and charge each revision for the plan distance it introduces. Figure~\ref{fig:regimes} maps the three commitment regimes in $(\cP, \sigma)$. Falling replanning costs move a deployment to the left, into the region where $\sigma$ alone determines its regime.

\section{Repeated disruptions: hysteresis and patch debt}\label{sec:dynamic}

A static model cannot represent the behavior most characteristic of commitment, which is continuing to execute a plan one would not choose fresh. We now let disruptions accumulate and revisions recur.

Time is discrete, with discount factor $\beta \in (0,1)$. The state is $(x, y)$. The first coordinate, $x \in [0, \bar{x}]$, is accumulated mismatch between plan and environment: the plan remains executable but incurs a flow cost $\theta(x)$ per period, with $\theta$ increasing and $\theta(0) = 0$. The second, $y \in \{0, 1, \dots\}$, is \emph{patch debt}, the number of repairs since the last fresh plan; patch debt incurs a flow cost $\Gamma(y)$. That $\Gamma$ is monotone is a modeling assumption. It encodes the intuition that each repair is a constrained search near an already-patched object, the planning analogue of technical debt in software. Repairs that refactor the plan can violate the assumption (Section~\ref{sec:limits}). Each period a shock arrives with probability $p$ and adds $\varepsilon$ to $x$. Before executing, the agent may \emph{continue}; \emph{repair} (pay $\kappa_{\mathrm{R}} + \cR(x) + \sigma \dR(x) n$, reset $x \to 0$ and $y \to y+1$); \emph{replan} (pay $\kappa_{\mathrm{P}} + \sigma \dP n$, reset $x \to 0$ and $y \to 0$); or \emph{abandon} (collect a terminal payoff $W$, which, unlike in Section~\ref{sec:model}, may be negative).

\begin{proposition}[Hysteresis and patch-then-rebuild]\label{prop:hysteresis}
Assume $\theta, \Gamma, \cR, \dR$ are continuous, nondecreasing, and bounded, and $\beta \in (0,1)$. Then:
\begin{enumerate}
\item[(i)] A unique bounded value function $V(x,y)$ exists and is nonincreasing in $x$ and in $y$.
\item[(ii)] \emph{(Inaction band.)} If all revision frictions vanish ($\kappa_{\mathrm{R}} = \kappa_{\mathrm{P}} = \sigma = 0$, $\cR \equiv 0$), revision weakly dominates continuing at every $x > 0$. If instead every revision costs at least $\underline{\kappa} > 0$ and abandonment is not optimal at $(0,0)$, the continue region contains an interval $[0, x_{\ell}]$ with $x_{\ell} > 0$. This is commitment hysteresis: the agent keeps executing plans that are strictly worse than a free fresh start would deliver.
\item[(iii)] \emph{(Threshold structure.)} If $x \mapsto \theta(x) - \cR(x) - \sigma n\, \dR(x)$ is nondecreasing (mismatch costs grow at least as fast as repair's cost penalty), then for each $y$ the continue region is an interval $[0, \bar{x}(y))$: revise exactly when mismatch crosses a threshold.
\item[(iv)] \emph{(Patch-then-rebuild.)} Let $\Gamma$ be strictly increasing. At any revision moment, the preference for repair over replanning is strictly decreasing in $y$: there is $\bar{y}(x) \in \mathbb{N} \cup \{\infty\}$ such that repair is chosen when $y < \bar{y}(x)$ and replanning when $y \ge \bar{y}(x)$. If, moreover, the debt burden eventually exceeds replanning's extra cost, that is, $\lim_{y}\Gamma(y) - \Gamma(0) > \kappa_{\mathrm{P}} + \sigma n \dP - \kappa_{\mathrm{R}}$, then $\bar{y}(x)$ is finite at every revision state; along any optimal path that keeps revising, each run of consecutive repairs is finite and ends in a rebuild. With bounded debt costs below that gap, patching forever can be optimal.
\end{enumerate}
\end{proposition}

\begin{figure}[t]
\centering
\includegraphics[width=0.72\linewidth]{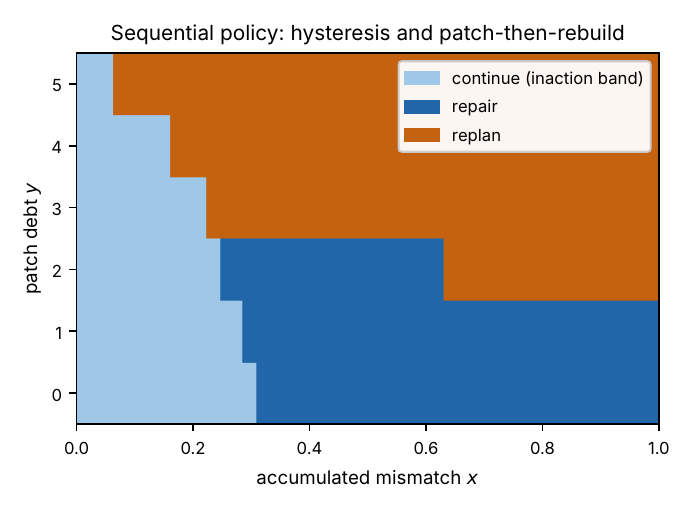}
\caption{Optimal sequential policy in (mismatch $x$, patch debt $y$), computed by value iteration at the parameters listed in Section~\ref{sec:numerics}. Light blue: continue (the hysteresis band), which narrows as $y$ grows; dark blue: repair; orange: replan. Repairs accumulate patch debt until rebuilding dominates (Proposition~\ref{prop:hysteresis}(iv)).}
\label{fig:hysteresis}
\end{figure}

Figure~\ref{fig:hysteresis} shows the computed policy. The inaction band narrows as patch debt grows, since a much-patched plan merits less tolerance. Beyond the patch-debt threshold, replanning replaces repair, and at intermediate debt the static depth threshold of Proposition~\ref{prop:threshold} reappears as a boundary in $x$ between repair and replanning.

In a two-period version of the model, the comparative statics of the band width hold in closed form (Appendix~\ref{app:twoperiod}). In the infinite-horizon model they are not proved, and we check them numerically: raising the revision frictions $\kappa$ never narrowed the band ($25$ of $25$ parameter draws; strictly wider in $23$), and neither did raising the stability price $\sigma$ ($25$ of $25$; strictly wider in $17$). The scans also showed that in $22$ of $25$ draws the band strictly widened with the shock probability $p$: when disruptions are frequent, a repaired plan is itself short-lived, so prompt repair is worth less. We report this as a numerical regularity with an intuitive mechanism and do not claim it as a theorem for the infinite-horizon model. In the three exceptions, two draws showed no change and one showed a narrow reversal near the abandonment boundary (Table~\ref{tab:verification}).

\section{The value of diagnosis}\label{sec:voi}

The policies above take the depth $d$ as known. Measuring $d$ takes work, such as validating the plan against the new state, running simulations, or probing tools, at a compute cost $\cdiag$. (Stopping rules for verify--repair loops in LLM agents address the same question at the systems level \cite{vrr2026}.) Suppose the agent knows that a disruption has occurred but not its depth, and holds a prior $F$ on $[0,1]$. Let $U_{\mathrm{fb}} = \max\{\up, \ua\}$ be the best fallback. Acting uninformed yields $\max\{\mathbb{E}_F\, \ur(d),\; U_{\mathrm{fb}}\}$; diagnosing first yields $\mathbb{E}_F \max\{\ur(d), U_{\mathrm{fb}}\} - \cdiag$. The value of information in the classical sense \cite{howard1966}, applied to the revision decision, is
\[
\mathrm{VoI}(F) \;=\; \mathbb{E}_F \max\{\ur(d), U_{\mathrm{fb}}\} \;-\; \max\{\mathbb{E}_F\, \ur(d),\; U_{\mathrm{fb}}\}.
\]

\begin{proposition}[When to measure the damage]\label{prop:voi}
\begin{enumerate}
\item[(i)] $\mathrm{VoI}(F) \ge 0$ always, and $\mathrm{VoI}(F) = 0$ if and only if a single action is optimal $F$-almost surely. In particular, when $U_{\mathrm{fb}} = \up$ and $F(\dstar) = 0$, diagnosis is worthless and ``always replan'' is optimal: environments with predictably deep damage rationalize the convention.
\item[(ii)] Diagnosis is strictly optimal if $\cdiag < \mathrm{VoI}(F)$ and strictly suboptimal if $\cdiag > \mathrm{VoI}(F)$, with indifference at equality.
\item[(iii)] Write $\varphi_{\mathrm{fb}}(d) = \ur(d) - U_{\mathrm{fb}}$, with threshold $d^{*}_{\mathrm{fb}}$ (equal to $\dstar$ when $U_{\mathrm{fb}} = \up$ and to $d^{\circ}$ when $U_{\mathrm{fb}} = \ua$). For symmetric two-point priors at $d^{*}_{\mathrm{fb}} \pm s$, $\mathrm{VoI}(s) = \tfrac{1}{2}\min\{\varphi_{\mathrm{fb}}(d^{*}_{\mathrm{fb}} - s),\, -\varphi_{\mathrm{fb}}(d^{*}_{\mathrm{fb}} + s)\}$, which is nondecreasing in the spread $s$, and strictly increasing while both points remain interior to $(0,1)$.
\end{enumerate}
\end{proposition}

\begin{figure}[t]
\centering
\includegraphics[width=0.72\linewidth]{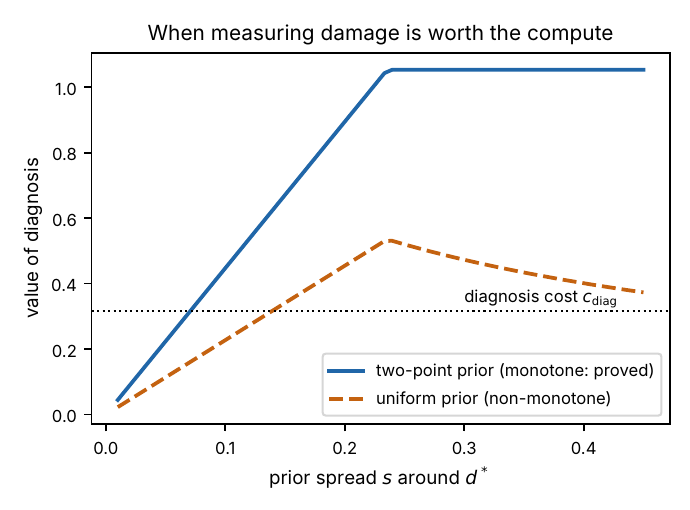}
\caption{Value of diagnosis against prior spread $s$ around $\dstar$ for two-point priors (solid; monotone by Proposition~\ref{prop:voi}(iii)) and uniform priors (dashed), at a parameterization selected by the checker. The dashed curve turns down where the widening support is truncated at the boundary of $[0,1]$ (Remark~\ref{rem:uniform}); for interior spreads the uniform curve is monotone. Diagnosis pays when its value exceeds $\cdiag$ (dotted line).}
\label{fig:voi}
\end{figure}

\needspace{13\baselineskip}
\begin{remark}[Where non-monotonicity comes from]\label{rem:uniform}
A wider prior does not always make diagnosis more valuable. For uniform priors centered at the threshold, $\mathrm{VoI}$ is nondecreasing in the spread \emph{as long as the support stays inside} $[0,1]$, because each branch of the min averages a monotone integrand (verified on $400$ instances with no exceptions). The rise and fall of the dashed curve in Figure~\ref{fig:voi} occurs where the widening support is truncated at the boundary of $[0,1]$. The prior is then no longer symmetric about the threshold, and the added mass falls where the uninformed action is already correct; this non-monotonicity appears in $375$ of $400$ truncated instances. That the value of information need not increase with prior uncertainty is a classical point \cite{gould1974,hilton1981}. The contribution here is its threshold form: diagnosis has value only when the prior places mass on both sides of the revision threshold, and the size of that value depends on where the mass lies, which total uncertainty does not measure.
\end{remark}

\section{Numerical verification and calibration}\label{sec:numerics}

\paragraph{Verification protocol.}
A separate checker implements only the primitives \eqref{eq:UR}--\eqref{eq:UA} and \eqref{eq:cone} and the sequential Bellman operator, and tests each proposition's claims against them. It has five components.
\begin{enumerate}
\item \emph{Symbolic audit.} For a linear family, computer algebra confirms eight properties of the closed forms of Propositions~\ref{prop:threshold} and~\ref{prop:sigma}, including the closed-form derivative of $\dstar$ in $\sigma$ and the signs of the comparative statics in $\cP$ and $\Delta$.
\item \emph{Static scan.} Of $4{,}000$ randomly drawn instances with piecewise-linear increasing primitives (the scans draw $n$ continuously, which the payoffs read only through $\sigma n$), the $3{,}946$ that satisfy the boundary condition of Assumption~\ref{ass:a1} were evaluated on a $401$-point depth grid ($1.6$ million evaluations). There were zero violations of monotonicity, sign, uniqueness, or the comparative statics in $\sigma$. Both regimes occurred ($2{,}891$ interior and $1{,}055$ lock-in instances), and in $1{,}475$ instances abandonment was the operative fallback. Corollary~\ref{cor:scale} held in all $395$ instances on which it was tested, $294$ of which satisfy Assumption~\ref{ass:a1} at both sizes. A separate scan of $3{,}959$ admissible instances of the revision-risk extension (Appendix~\ref{app:general}) found at most one sign change of $\tilde\varphi$ in every instance.
\item \emph{Graph enumeration.} Exact reachability enumeration on $600$ random DAGs confirms \eqref{eq:alpha} to machine precision, together with the bounds, the extremal characterizations, and the parallel-chain formula. A separate search of $600$ DAGs found no violation of the serial maximum.
\item \emph{Sequential model.} Value iteration on a $61 \times 9$ grid solves the model for $59$ random parameter draws. All $59$ optimal policies have threshold structure, and patch-then-rebuild cycles appear in all $26$ draws whose optimal policy uses both instruments. The band statics are those reported in Section~\ref{sec:dynamic}. A control without frictions shrinks the band to the grid's resolution. In all $16$ bounded-debt draws that satisfy the finiteness condition of Proposition~\ref{prop:hysteresis}(iv), the optimal policy uses replanning.
\item \emph{Value of information.} Across $3{,}000$ instances, nonnegativity is never violated, VoI is zero in all $626$ instances whose prior lies entirely above the threshold, and two-point monotonicity holds in all $140$ draws tested. A dedicated experiment isolates the truncation mechanism of Remark~\ref{rem:uniform}: interior spreads are monotone in $400$ of $400$ instances, and truncated spreads are non-monotone in $375$ of $400$.
\end{enumerate}
Numerical tolerances were $10^{-9}$ for the scans and $10^{-12}$ for the enumeration, and no claimed result failed a check. Table~\ref{tab:verification} summarizes the results.

\begin{table}[t]
\centering\footnotesize
\setlength{\tabcolsep}{3.5pt}
\begin{tabular}{@{}llll@{}}
\toprule
Component & Claims tested & Scale & Outcome \\
\midrule
Symbolic audit & Props.~\ref{prop:threshold} and~\ref{prop:sigma} closed forms & 8 checks & all pass \\
Static scan & monotonicity, uniqueness, regimes, $\sigma$-statics & 3{,}946 inst.\ $\times$ 401 grid & 0 violations \\
Scale corollary & Corollary~\ref{cor:scale} & 395 (294 admissible) & 0 violations \\
Revision risk & App.~\ref{app:general} single crossing & 3{,}959 instances & 0 violations \\
Graph enumeration & \eqref{eq:alpha}, bounds, extremes, parallel & 600 random DAGs & exact ($10^{-12}$) \\
Sequential DP & threshold structure; patch-then-rebuild & 59 draws ($61 \times 9$ grid) & 59/59; 26/26 \\
Band statics & $\kappa$, $\sigma$: no narrowing; $p$: widening & 25 draws each & 25/25; 25/25; 22/25$^{\dagger}$ \\
Finiteness condition & Prop.~\ref{prop:hysteresis}(iv) & 16 bounded-debt draws & 16/16 \\
VoI & nonneg.; zero above $\dstar$; two-point monotone & 3{,}000 / 626 / 140 & 0 viol.; 626/626; 140/140 \\
Truncation mechanism & Remark~\ref{rem:uniform} & 400 interior / 400 truncated & 400/400; 375/400$^{\ddagger}$ \\
\bottomrule
\end{tabular}
\caption{Verification summary. $^{\dagger}$Numerical regularity: 2 flat, 1 reversal near the abandonment boundary. $^{\ddagger}$Interior spreads: monotone in 400 of 400. Truncated spreads: non-monotone in 375 of 400.}
\label{tab:verification}
\end{table}

\begin{figure}[t]
\centering
\includegraphics[width=0.78\linewidth]{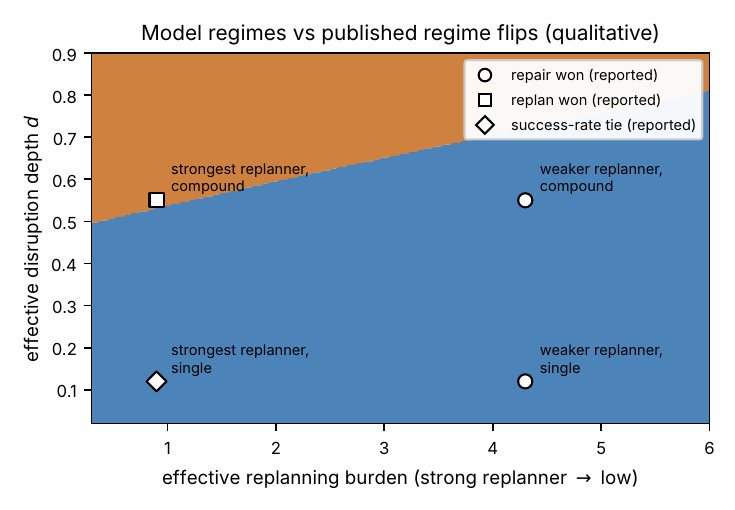}
\caption{Model policy regions (blue: repair; orange: replan) in (effective replanning burden $\cP + (1 - q_{\mathrm{P}})(V - W - \sigma \dP n)$, depth $d$), with the four experimental cells of \cite{yuan2026replan} placed qualitatively. Circles mark cells where the authors report hierarchical repair winning; the square marks the cell where replanning won (compound disruptions, strongest replanner); the diamond marks the success-rate tie, a cell uninformative on success rate alone and consistent with the model only through the stability term ($\sigma > 0$). The placements are illustrative: stronger reported replanners are placed at lower burden (through $q_{\mathrm{P}}$) and compound disruptions are placed at higher depth; no parameters were fit.}
\label{fig:calibration}
\end{figure}

\paragraph{Figure parameters.}
Figures~\ref{fig:policy}, \ref{fig:regimes}, and~\ref{fig:calibration} use the baseline $n = 30$, $V = 40$, $\cP = 3$, $\dP = 0.8$, $\cR(d) = 6d^{1.2}$, $\Delta(d) = 8d^{1.5}$, and $\dR(d) = 0.25d$, with $W = 30$ in Figure~\ref{fig:policy} and $W = 20$ elsewhere. Figure~\ref{fig:topology} draws $300$ DAGs, each with $n$ uniform on $\{4, \dots, 60\}$ and each forward edge present independently with a probability drawn uniformly from $[0.02, 0.6]$. Figure~\ref{fig:regimes} varies $\cP \in [0.01, 8]$ and $\sigma \in [0, 1.2]$; Figure~\ref{fig:calibration} fixes $\sigma = 0.25$ and varies the effective replanning burden over $[0.3, 6]$. Figure~\ref{fig:hysteresis} solves the sequential model on an $81 \times 6$ grid (debt truncated at $y = 5$) with $\beta = 0.95$, $p = 0.25$, $\varepsilon = 0.2$, $\bar{x} = 1$, $\theta(x) = 2x^{1.5}$, $\Gamma(y) = 0.09y$, $\cR(x) = 0.6x$, $\dR(x) = 0.15x$, $\kappa_{\mathrm{R}} = 0.35$, $\kappa_{\mathrm{P}} = 3.2$, $\sigma = 0.25$, $n = 3$, $\dP = 0.8$, and $W = -30$. Figure~\ref{fig:voi} uses the parameterization found by the checker's counterexample search: $V = 68.5$, $W = 7.88$, $\sigma n = 0.865$, $\cP = 1.62$, $\dP = 0.568$, $\cR(d) = 0.788d$, $\Delta(d) = 8.10d$, and $\dR(d) = 0.0684d$, with $\cdiag$ drawn at $30\%$ of the two-point maximum; the payoffs \eqref{eq:UR}--\eqref{eq:UA} read $n$ and $\sigma$ only through the product $\sigma n$.

\paragraph{Qualitative calibration.}
Two published experimental records serve as external anchors, and we cite their findings as their authors report them. In the DriverLog-Time experiments of Fox et~al.\ \cite{fox2006}, repair changed about one-seventh as many of the original plan's first-step actions as replanning did ($8.9\%$ versus $66.0\%$); the authors also report that as little as 15--20\% of the original plan survives replanning, and across their domains repair's advantage grew with problem size. These findings correspond to the model's premise $\dR \ll \dP$ and to the prediction of Corollary~\ref{cor:scale}. Across the three revision strategies and systematic disruptions of Yuan et~al.\ \cite{yuan2026replan}, repair-style methods preserved $86$--$92\%$ of itinerary content versus $57$--$65\%$ for replanning, and hierarchical repair won or tied on success everywhere except under compound (deep) disruptions handled by the strongest replanning model, where full replanning pulled ahead. Figure~\ref{fig:calibration} locates that regime flip in the model's space of effective replanning burden and depth. A weak replanner fails with positive probability, which the reduction in Appendix~\ref{app:general} prices as an effective replanning burden $\cP + (1 - q_{\mathrm{P}})(V - W - \sigma \dP n)$. A stronger replanner therefore corresponds to a lower burden through $q_{\mathrm{P}}$ as well as through compute cost, and a compound disruption corresponds to a higher $d$. The single-disruption cell with the strongest replanner is a tie in success rate, which the authors themselves decline to rank, so success rate alone does not discriminate there. The same experiments show repair to be six times more stable in that cell, and the model prices that margin through $\sigma > 0$. Table~\ref{tab:calibration} collects these correspondences; no model parameter was fit to either dataset.

\begin{table}[t]
\centering\small
\begin{tabular}{@{}>{\raggedright\arraybackslash}p{0.47\linewidth}>{\raggedright\arraybackslash}p{0.47\linewidth}@{}}
\toprule
Reported finding (authors' own) & Model counterpart \\
\midrule
Repair changed $8.9\%$ of the original first-step actions; replanning $66.0\%$ (DriverLog-Time) \cite{fox2006} & the $\dR(d) \ll \dP$ premise \\ \addlinespace[3pt]
Repair's advantage grows with problem size \cite{fox2006} & Corollary~\ref{cor:scale} (scale widens the repair region) \\ \addlinespace[3pt]
Repair preserved $86$--$92\%$ of itinerary content vs $57$--$65\%$ for replanning \cite{yuan2026replan} & the $\dR(d) \ll \dP$ premise, second domain \\ \addlinespace[3pt]
Replanning beat hierarchical repair only at compound disruptions with the strongest replanner \cite{yuan2026replan} & regime flip at high $d$, low effective replanning burden (Figure~\ref{fig:calibration}) \\ \addlinespace[3pt]
Success-rate tie at single disruptions, strongest replanner, with repair $6\times$ more stable \cite{yuan2026replan} & cell uninformative on success alone; decided by the stability term ($\sigma > 0$) \\ \addlinespace[3pt]
No fixed strategy dominates \cite{yuan2026replan} & parameter-dependent regime taxonomy (Proposition~\ref{prop:threshold}) \\
\bottomrule
\end{tabular}
\caption{Correspondence between published experimental findings and model counterparts. The correspondence is qualitative; the findings are those reported by the cited authors; the experiments were not rerun here.}
\label{tab:calibration}
\end{table}

\needspace{14\baselineskip}
\section{Implications for agent design}\label{sec:implications}

\paragraph{Replace revision conventions with a measured threshold.}
The quantities the policy needs can all be measured or assigned in a deployed agent framework. Some are logged or computed directly: raw damage from failed precondition checks, amplification from the plan's own dependency graph \eqref{eq:alpha}, repair and replanning cost from token and latency logs, and plan distance from the action-set metric of \cite{fox2006}. Others must be estimated or assigned. The quality gap $\Delta$ is observable only through occasional paired audits that run both instruments on the same disruption; the success probabilities $q_{\mathrm{R}}$ and $q_{\mathrm{P}}$ must be estimated; and the stability price $\sigma$ is a policy choice that encodes what plan churn costs the deployment. A framework that logs the first group and estimates or sets the second explicitly can implement Proposition~\ref{prop:threshold} directly.

\paragraph{Price plan churn before planning gets cheaper.}
By Corollary~\ref{cor:cheap}, as planners become cheaper, architectures that leave $\sigma$ at zero will move into the churn regime, where agents discard and rebuild their plans at nearly every disturbance. Each such revision is optimal for the agent's own objective, yet it imposes costs that the objective omits wherever humans approve plans, other agents coordinate with the agent, or audits require stable intentions. The design response is to charge each revision explicitly for the plan distance it introduces, at a rate set by the deployment's actual coordination costs.

\needspace{10\baselineskip}
\paragraph{Structure plans for repairability.}
Proposition~\ref{prop:topology} turns a vague preference for ``modular'' plans into a quantitative exchange rate. The expected validation footprint grows with $R(G)$, and replacing a serial plan of $n$ actions with $w$ parallel chains of equal length divides $\alpha$ by $w(n+1)/(n+w)$, a factor that approaches $w$ for long plans. To first order, the tolerable raw damage rises by the same factor when broken actions have no effect-equivalent substitutes. Where the goal allows it, planners should be rewarded for producing plans with low $\alpha$, a statistic that can be computed from the plan graph before execution.

\paragraph{Budget diagnosis by threshold proximity.}
Proposition~\ref{prop:voi} implies that spending on diagnosis should be concentrated on disruptions whose damage prior straddles the threshold. Agent fleets that face predictably deep disruptions, and for which replanning beats abandoning, can hardwire replanning without loss. Fleets that face heterogeneous damage should spend compute measuring depth before revising.

\paragraph{Expect and manage hysteresis.}
Proposition~\ref{prop:hysteresis} reinterprets the commitment rules proposed against plan churn \cite{irrational2026}. When revisions carry frictions, an inaction band is the optimal response, and its width should be chosen deliberately through $\kappa$ and $\sigma$. Patch-debt accounting gives long-running agent plans a maintenance rule: track the repairs made since the last rebuild, and rebuild when the debt threshold is crossed.

\section{Limitations}\label{sec:limits}

The model prices revision instruments in reduced form. It does not derive $\cR$, $\Delta$, or $\dR$ from a specific repair algorithm, and real repair systems vary \cite{gerevini2003,saetti2025}. The footprint $I(B)$ counts every causally downstream action as damage, which is exact when broken actions lack effect-equivalent substitutes and an overestimate otherwise; the topology results should be read as worst-case validation bounds. The stability price $\sigma$ aggregates coordination, oversight, and commitment costs without modeling the strategic behavior that generates them, and its linearity in distance is an approximation (Remark~\ref{rem:tariff} treats the lumpy case). Endogenizing $\sigma$ requires a multi-agent model and is beyond the scope of this paper. Monotone patch debt $\Gamma$ is an assumption: repairs that refactor the plan can reduce debt and break the single crossing in Proposition~\ref{prop:hysteresis}(iv). The amplification law is exact for a single uniformly random broken action and gives only an upper bound for multi-action damage. Proposition~\ref{prop:hysteresis}(iii) relies on a sufficient condition relating mismatch costs to repair costs. Outside that condition, non-threshold policies are possible in principle, although none appeared in the $59$ randomized solutions. The calibration in Section~\ref{sec:numerics} is a qualitative correspondence with two published experimental records, and no parameters were estimated. All numerical results are model computations; no new experiments with deployed agents were run. Finally, the theory optimizes a single agent's objective. Whether churn that is optimal for each agent is also optimal for a population of interacting agents is an open question, which the stability price addresses only in reduced form.

\section*{Disclosure}

The author is affiliated with Omega Venture Partners, an investment firm focused on artificial intelligence, and may hold investments whose value could be affected by the themes discussed. This is a theoretical paper; its numerical results are model computations, not empirical estimates, forecasts, or investment recommendations. The paper uses no confidential company, portfolio, customer, founder, diligence, or transaction information. No external funding was received. Generative AI tools assisted with language editing, typesetting, and implementation of numerical checks. The author verified all analytical and numerical results and the final text.

\section*{Data and code availability}

The paper contains no proprietary or confidential data. The propositions and corollaries are proved in Appendices~\ref{app:proofs}--\ref{app:twoperiod}. The figures and the verification results in Table~\ref{tab:verification} were generated by the model computations reported in Section~\ref{sec:numerics}, which also lists the figure parameters. The experimental cells in Figure~\ref{fig:calibration} are placed qualitatively, and Table~\ref{tab:calibration} reports findings of the cited authors. No separate data or computational package accompanies this version.

\appendix

\section{Proofs}\label{app:proofs}

\subsection{Proof of Proposition~\ref{prop:threshold}}

\emph{(i)} $\varphi(d) = [\cP - \cR(d)] - \Delta(d) + \sigma n[\dP - \dR(d)]$ is a sum of strictly decreasing functions ($\cR, \Delta, \dR$ strictly increasing; $\sigma n \ge 0$) plus constants, hence strictly decreasing; $\varphi(0) = \cP + \sigma n \dP > 0$ by Assumption~\ref{ass:a1}. $\varphi(1) = [\cP - \cR(1) - \Delta(1)] + \sigma n [\dP - \dR(1)]$. If $\dR(1) < \dP$: $\varphi(1) < 0 \iff \sigma < \sigma_{\mathrm{lock}}$ by \eqref{eq:lock} and $\cR(1) + \Delta(1) \ge \cP$. Strict monotonicity, continuity, and the intermediate value theorem give a unique root $\dstar \in (0,1]$, with $\varphi > 0$ below and $\varphi < 0$ above.

\emph{(ii)} If $\sigma \ge \sigma_{\mathrm{lock}}$ then $\varphi(1) \ge 0$; by monotonicity $\varphi(d) \ge \varphi(1) \ge 0$ for all $d \le 1$. The case split in \eqref{eq:lock} covers the boundary configurations. When $\dR(1) = \dP$ and $\cR(1) + \Delta(1) > \cP$, $\varphi(1) = \cP - \cR(1) - \Delta(1) < 0$ for \emph{every} $\sigma$, so lock-in is unattainable, and $\sigma_{\mathrm{lock}} = +\infty$ encodes that the interior regime holds at all $\sigma$. When $\dR(1) = \dP$ and $\cR(1)+\Delta(1) = \cP$, $\varphi(1)=0$ identically, so every $\sigma$ sits on the lock-in boundary (repair weakly dominant everywhere, indifference at $d=1$), which $\sigma_{\mathrm{lock}} = 0$ encodes.

\emph{(iii)} If $\up \ge W$: for $d < \dstar$, $\ur(d) > \up \ge \ua$; for $d > \dstar$, $\up > \ur(d)$ and $\up \ge \ua$. If $\up < W$: replanning is dominated by abandonment everywhere; $\ur(0) = V$ and $\ur$ is strictly decreasing, so repair beats abandonment iff $\ur(d) > W$, i.e.\ $d < d^{\circ}$, where $d^{\circ}$ is the unique root of $\ur(d) = W$ when $V > W \ge \ur(1)$; if $W \ge V$, abandon everywhere; if $\ur(1) > W$, repair everywhere.

\emph{(iv)} For $\sigma' > \sigma$: $\varphi_{\sigma'}(d) - \varphi_{\sigma}(d) = (\sigma' - \sigma) n [\dP - \dR(d)] \ge 0$, with strict inequality wherever $\dR(d) < \dP$. Since $\dR(\dstar) \le \dR(1) \le \dP$ and the inequality is strict on the interior regime (otherwise we would be in case (ii)), $\varphi_{\sigma'}(\dstar_{\sigma}) > 0$, and strict monotonicity of $\varphi_{\sigma'}$ places its root strictly above $\dstar_{\sigma}$. The same pointwise-shift argument applies verbatim to increases in $\cP$ (shift $+\mathrm{d}\cP > 0$) and to pointwise upward shifts of $\Delta$, $\cR$, $\sigma n\,\dR$ (downward shifts of $\varphi$). $\qed$

\subsection{Proof of Corollary~\ref{cor:scale}}

Write $\varphi_n(d) = \kappa[C(n) - h(dn)] - \Delta(d) + \sigma n [\dP - \dR(d)]$. The stated condition says exactly that $\varphi_{n'}(d) \ge \varphi_{n}(d)$ for every interior $d$ when $n' > n$. If $\dstar_n = 1$ (lock-in), the convention gives $\dstar_{n'} \ge \dstar_n$ only if lock-in persists, which follows because $\varphi_{n'}(1) \ge \varphi_n(1) \ge 0$ whenever the condition extends to $d = 1$ by continuity. Otherwise $\varphi_{n'}(\dstar_n) \ge \varphi_{n}(\dstar_n) = 0$, and since $\varphi_{n'}$ is strictly decreasing in $d$ (Assumption~\ref{ass:a1} applies at every admissible $n$), its root is $\ge \dstar_n$, strictly if the inequality is strict at $\dstar_n$; under lock-in at both sizes, $\dstar_{n'} = \dstar_n = 1$ and strictness is vacuous. $\qed$

\subsection{Proof of Proposition~\ref{prop:topology} and Corollary~\ref{cor:serial}}

For single uniform $B = \{a\}$, $|I(\{a\})| = |\{a\} \cup \{v : a \leadsto v\}| = 1 + r(a)$, where $r(a)$ is the number of actions reachable from $a$. Hence $\mathbb{E}|I| = 1 + \frac{1}{n}\sum_{a} r(a) = 1 + R(G)/n$, which is \eqref{eq:alpha}. Each unordered pair contributes at most one ordered reachability pair in a DAG, since $u \leadsto v$ and $v \leadsto u$ together would create a cycle; hence $0 \le R(G) \le \binom{n}{2} = n(n-1)/2$ and $1 \le \alpha \le 1 + (n-1)/2 = (n+1)/2$. Equality below holds iff $R(G) = 0$ iff $E = \emptyset$. Equality above holds iff every pair is comparable under reachability, i.e.\ the transitive closure is a total order; the chain attains it. For $w$ parallel chains of length $L = n/w$, a node at depth $i$ in its chain ($i = 1, \dots, L$) reaches $L - i$ others, so $R = w \sum_{i=1}^{L} (L - i) = w L(L-1)/2$, and with $n = wL$, $\alpha = 1 + \frac{wL(L-1)/2}{wL} = 1 + \frac{L-1}{2} = \frac{L+1}{2}$. For general $B$, $I(B) = \bigcup_{a \in B} I(\{a\})$, so $|I(B)| \le \min\{n, \sum_{a \in B} |I(\{a\})|\}$ pointwise, and the bound passes to expectations. For Corollary~\ref{cor:serial}: with $d \approx \alpha b$, depth reaches a threshold $\bar d$ at raw damage $b \approx \bar d/\alpha$, and the serial and $w$-chain values $\alpha = (n+1)/2$ and $\alpha = (n/w+1)/2$ give the ratio $w(n+1)/(n+w)$, which approaches $w$ as $n$ grows. $\qed$

\subsection{Proof of Proposition~\ref{prop:sigma} and Corollary~\ref{cor:cheap}}

Repair is optimal against replanning iff $\varphi(d) \ge 0$ iff $\sigma n [\dP - \dR(d)] \ge \Delta(d) + \cR(d) - \cP$. When the right side is positive and $\dR(d) < \dP$, dividing gives $\sigma \ge \sigma^{*}(d)$ as in \eqref{eq:sigmastar}. The limit as $\cP \to 0$ is immediate; positivity for $d > 0$ follows from $\Delta$ and $\cR$ being strictly increasing from $0$. For the churn limit: with $\sigma = 0$, $\varphi(d) = \cP - \cR(d) - \Delta(d)$, whose root $\dstar(\cP)$ satisfies $\cR(\dstar) + \Delta(\dstar) = \cP \to 0$, forcing $\dstar \to 0$ by strict monotonicity and continuity at $0$. $\qed$

\subsection{Proof of Proposition~\ref{prop:hysteresis}}\label{app:hysteresis}

\emph{(i)} The action set is finite, per-period rewards are bounded (all primitives are bounded on the compact state space and $W$ is finite), and $\beta < 1$. Consider the Bellman operator
\[
(TV)(x,y) = \max\{\,V_{\mathrm{cont}},\, V_{\mathrm{rep}},\, V_{\mathrm{replan}},\, W\,\}
\]
with
$V_{\mathrm{cont}} = -\theta(x) - \Gamma(y) + \beta \mathbb{E}_x V$,
$V_{\mathrm{rep}} = -\kappa_{\mathrm{R}} - \cR(x) - \sigma n \dR(x) - \theta(0) - \Gamma(y{+}1) + \beta \mathbb{E}_0 V(\cdot, y{+}1)$,
$V_{\mathrm{replan}} = -\kappa_{\mathrm{P}} - \sigma n \dP - \theta(0) - \Gamma(0) + \beta \mathbb{E}_0 V(\cdot, 0)$,
where $\mathbb{E}_x V = (1-p) V(x, \cdot) + p V(\min(x + \varepsilon, \bar x), \cdot)$.
This operator is a $\beta$-contraction on bounded functions, so a unique bounded fixed point exists. Monotonicity: if $V$ is nonincreasing in $x$ and $y$, so is $TV$. $V_{\mathrm{cont}}$ inherits the property from $-\theta$, $-\Gamma$, the monotone transition $x \mapsto \min(x+\varepsilon, \bar x)$, and $V$; $V_{\mathrm{rep}}$ is nonincreasing in $x$ (through $-\cR - \sigma n \dR$) and in $y$ (through $-\Gamma(y{+}1)$ and $V(\cdot, y{+}1)$); $V_{\mathrm{replan}}$ and $W$ are constant in $(x,y)$; and the maximum of nonincreasing functions is nonincreasing. Because the property is preserved under uniform limits, the fixed point inherits it.

\emph{(ii)} \emph{Frictionless case.} With $\kappa_{\mathrm{R}} = \kappa_{\mathrm{P}} = \sigma = 0$ and $\cR \equiv 0$, compare continuing with replanning, which is then free and resets both coordinates:
$V_{\mathrm{replan}}(y) - V_{\mathrm{cont}}(x,y) = [\theta(x) - \theta(0)] + [\Gamma(y) - \Gamma(0)] + \beta\,[\mathbb{E}_0 V(\cdot, 0) - \mathbb{E}_x V(\cdot, y)] \;\ge\; 0$,
each bracket being nonnegative by monotonicity of $\theta$, $\Gamma$, the transition, and $V$ (part (i)). Hence revision weakly dominates continuing at every state, strictly whenever $\theta(x) > \theta(0)$ or $\Gamma(y) > \Gamma(0)$, and the inaction band is degenerate. \emph{Frictional case.} Suppose every revision costs at least $\underline{\kappa} > 0$. At $(0,0)$, $V_{\mathrm{replan}}(0) = -\underline{\kappa}' - \theta(0) - \Gamma(0) + \beta \mathbb{E}_0 V(\cdot,0) \le V_{\mathrm{cont}}(0,0) - \underline{\kappa}$, with $\underline{\kappa}' \ge \underline{\kappa}$ the replanning friction, and $V_{\mathrm{rep}}(0,0) \le -\underline{\kappa} - \theta(0) - \Gamma(1) + \beta \mathbb{E}_0 V(\cdot,1) \le V_{\mathrm{cont}}(0,0) - \underline{\kappa}$, using $\Gamma(1) \ge \Gamma(0)$ and monotonicity of $V$ in $y$. Since abandonment is not optimal at $(0,0)$ and continuing beats both revisions there, $W < V_{\mathrm{cont}}(0,0)$. So continuing is strictly optimal at $(0,0)$, by a positive margin. By continuity of $\theta$, $\cR$, and $\dR$ in $x$ (all value differences above are continuous in $x$), the strict preference persists on an interval $[0, x_{\ell}]$ with $x_{\ell} > 0$.

\emph{(iii)} Fix $y$; let $g(x) = V_{\mathrm{cont}}(x,y) - \max\{V_{\mathrm{rep}}(x,y), V_{\mathrm{replan}}(y), W\}$, suppose $g(x_2) \le 0$ (revision optimal at $x_2$), and take $x \ge x_2$. Use $\max\{a', c\} - \max\{a, c\} \le \max\{a' - a, 0\}$ with $a = V_{\mathrm{rep}}(x,y)$, $a' = V_{\mathrm{rep}}(x_2,y)$, and $c = \max\{V_{\mathrm{replan}}, W\}$ (constant in $x$), where $a' - a = [\cR(x) + \sigma n \dR(x)] - [\cR(x_2) + \sigma n \dR(x_2)] \ge 0$. Then
\[
g(x) - g(x_2) \le [\theta(x_2) - \theta(x)] + \beta[\mathbb{E}_x V - \mathbb{E}_{x_2} V] + \bigl([\cR(x) + \sigma n \dR(x)] - [\cR(x_2) + \sigma n \dR(x_2)]\bigr).
\]
The middle term is $\le 0$ by monotonicity of $V$ and of the transition. The first and third terms sum to $-\bigl[(\theta - \cR - \sigma n \dR)(x) - (\theta - \cR - \sigma n \dR)(x_2)\bigr] \le 0$ under the stated condition. Hence $g(x) \le g(x_2) \le 0$: the revision region is up-closed in $x$, so the continue region is an interval $[0, \bar x(y))$.

\emph{(iv)} At a revision moment, repair is preferred to replanning iff
$
V_{\mathrm{rep}} - V_{\mathrm{replan}} = \bigl[\kappa_{\mathrm{P}} + \sigma n \dP - \kappa_{\mathrm{R}} - \cR(x) - \sigma n \dR(x)\bigr] + \bigl[\Gamma(0) - \Gamma(y{+}1)\bigr] + \beta\bigl[\mathbb{E}_0 V(\cdot, y{+}1) - \mathbb{E}_0 V(\cdot, 0)\bigr] \ge 0.
$
The first bracket is constant in $y$; the second is strictly decreasing in $y$ ($\Gamma$ is strictly increasing, as hypothesized in (iv)); the third is nonincreasing in $y$ ($V$ is nonincreasing in $y$). The whole expression is strictly decreasing in $y$, so it changes sign at most once: repair for $y < \bar y(x)$ and replanning for $y \ge \bar y(x)$, with $\bar y(x) \in \mathbb{N} \cup \{\infty\}$.

\emph{Finiteness.} Using $\mathbb{E}_0 V(\cdot, y{+}1) \le \mathbb{E}_0 V(\cdot, 0)$, the display is bounded above by $G(x) - [\Gamma(y{+}1) - \Gamma(0)]$ with $G(x) = \kappa_{\mathrm{P}} + \sigma n \dP - \kappa_{\mathrm{R}} - \cR(x) - \sigma n \dR(x) \le \kappa_{\mathrm{P}} + \sigma n \dP - \kappa_{\mathrm{R}}$. If $\lim_{y}\Gamma(y) - \Gamma(0) > \kappa_{\mathrm{P}} + \sigma n \dP - \kappa_{\mathrm{R}}$, pick $y_{0}$ with $\Gamma(y_{0}{+}1) - \Gamma(0)$ exceeding that gap. The display is then negative for all $y \ge y_{0}$ at every $x$, so $\bar y(x) \le y_{0} < \infty$. The checker illustrates the condition on bounded-debt instances: all $16$ draws that satisfy it exhibit replanning in the optimal policy, and a bounded-debt instance that violates it exhibits repair at every debt level. $\qed$

\subsection{Proof of Proposition~\ref{prop:voi}}

\emph{(i)} The function $\max$ is convex, so $\mathrm{VoI} = \mathbb{E} \max\{\ur(d), U_{\mathrm{fb}}\} - \max\{\mathbb{E}\ur, U_{\mathrm{fb}}\} \ge 0$ by Jensen's inequality applied to the two-argument max with one constant argument. Equality analysis: if $\ur(d) \ge U_{\mathrm{fb}}$ $F$-a.s., both sides equal $\mathbb{E} \ur$; if $\ur(d) \le U_{\mathrm{fb}}$ $F$-a.s., both equal $U_{\mathrm{fb}}$. Conversely, if both events $\{\varphi_{\mathrm{fb}}(d) > 0\}$ and $\{\varphi_{\mathrm{fb}}(d) < 0\}$ (writing $\varphi_{\mathrm{fb}} = \ur - U_{\mathrm{fb}}$) have positive $F$-measure, then $\mathbb{E}\max > \max\{\mathbb{E}\ur, U_{\mathrm{fb}}\}$. Taking $\mathbb{E}\ur \ge U_{\mathrm{fb}}$ without loss of generality, $\mathbb{E}\max\{\ur, U_{\mathrm{fb}}\} = \mathbb{E}\ur + \mathbb{E}\max\{U_{\mathrm{fb}} - \ur, 0\} > \mathbb{E}\ur$, since the integrand is positive on a set of positive measure. The ``always replan'' special case is the instance $U_{\mathrm{fb}} = \up$, $F(\dstar) = 0$.

\emph{(ii)} Immediate from the definitions.

\emph{(iii)} With $d \in \{d^{*}_{\mathrm{fb}} - s, d^{*}_{\mathrm{fb}} + s\}$ equiprobable, $\varphi_{\mathrm{fb}}(d^{*}_{\mathrm{fb}} - s) \ge 0 \ge \varphi_{\mathrm{fb}}(d^{*}_{\mathrm{fb}} + s)$ by definition of the fallback-specific threshold (for $U_{\mathrm{fb}} = \up$, $\varphi_{\mathrm{fb}} = \varphi$ and $d^{*}_{\mathrm{fb}} = \dstar$; for $U_{\mathrm{fb}} = \ua$, $d^{*}_{\mathrm{fb}} = d^{\circ}$). Writing $\dstar$ for $d^{*}_{\mathrm{fb}}$ in what follows,
$\mathbb{E}\max\{\ur, U_{\mathrm{fb}}\} = U_{\mathrm{fb}} + \tfrac12 \varphi_{\mathrm{fb}}(\dstar{-}s)$ and $\max\{\mathbb{E}\ur, U_{\mathrm{fb}}\} = U_{\mathrm{fb}} + \max\{\tfrac12[\varphi_{\mathrm{fb}}(\dstar{-}s) + \varphi_{\mathrm{fb}}(\dstar{+}s)], 0\}$, so
\[
\mathrm{VoI}(s) = \tfrac12\,\varphi_{\mathrm{fb}}(\dstar{-}s) - \max\{\tfrac12[\varphi_{\mathrm{fb}}(\dstar{-}s) + \varphi_{\mathrm{fb}}(\dstar{+}s)],\, 0\} = \tfrac12 \min\{\varphi_{\mathrm{fb}}(\dstar{-}s),\, -\varphi_{\mathrm{fb}}(\dstar{+}s)\}.
\]
Both arguments of the min are nondecreasing in $s$ because $\varphi_{\mathrm{fb}}$ is strictly decreasing (Proposition~\ref{prop:threshold}(i), which applies to $\varphi_{\mathrm{fb}}$ with $U_{\mathrm{fb}}$ constant). The min of nondecreasing functions is nondecreasing, and it is strictly increasing while both points are interior to $(0,1)$. $\qed$

\section{General static model with revision risk}\label{app:general}

Let repair succeed with probability $q_{\mathrm{R}}(d) \in (0,1]$, nonincreasing in $d$, with failed repair falling back on replanning (its cost sunk), and let replanning succeed with probability $q_{\mathrm{P}} \in (0,1]$, with failure falling back on abandonment:
\begin{align*}
\tilde U_{\mathrm{P}} &= -\cP + q_{\mathrm{P}} [V - \sigma \dP n] + (1 - q_{\mathrm{P}}) W,\\
\tilde U_{\mathrm{R}}(d) &= -\cR(d) + q_{\mathrm{R}}(d) \bigl[V - \Delta(d) - \sigma \dR(d) n\bigr] + (1 - q_{\mathrm{R}}(d))\, \tilde U_{\mathrm{P}}.
\end{align*}
Then $\tilde\varphi(d) = \tilde U_{\mathrm{R}}(d) - \tilde U_{\mathrm{P}} = -\cR(d) + q_{\mathrm{R}}(d)\, g(d)$ with $g(d) = V - \Delta(d) - \sigma \dR(d) n - \tilde U_{\mathrm{P}}$, and $g$ is strictly decreasing. Augment Assumption~\ref{ass:a1} with $q_{\mathrm{R}}$ nonincreasing and $V > \tilde U_{\mathrm{P}}$ (a fresh plan beats the replanning lottery). On $\{d : g(d) \ge 0\}$, $\tilde\varphi$ is strictly decreasing, as the sum of a strictly decreasing and a nonincreasing term; on $\{d : g(d) < 0\}$, $\tilde\varphi(d) < 0$ regardless of monotonicity. Since $\tilde\varphi(0) = q_{\mathrm{R}}(0)[V - \tilde U_{\mathrm{P}}] > 0$ and $\{g \ge 0\}$ is an interval containing $0$, $\tilde\varphi$ changes sign at most once, from positive to negative. The threshold structure of Proposition~\ref{prop:threshold}(i) and (iii) therefore carries over with $\tilde\varphi$ in place of $\varphi$, single crossing replacing global monotonicity, and so do the comparative statics of part (iv) in $\cP$, $\Delta$, $\cR$, and $\dR$. The statics in $\sigma$ need an extra condition. Since $\partial\tilde\varphi/\partial\sigma = q_{\mathrm{R}}(d)\, n\,[q_{\mathrm{P}}\dP - \dR(d)]$, the threshold rises with $\sigma$ where $\dR < q_{\mathrm{P}}\dP$ at the threshold and can fall otherwise, because a failed replanning attempt incurs no distance charge. The lock-in boundary must be re-derived, because the boundary condition of Assumption~\ref{ass:a1} no longer determines the sign of $\tilde\varphi(1)$, and we do not claim \eqref{eq:lock} for this extension. A randomized scan of $3{,}959$ admissible instances on a $401$-point depth grid found $\tilde\varphi(0) > 0$ and at most one sign change in every instance.

\section{Two-period band statics}\label{app:twoperiod}

Consider two periods, no discounting, and $\theta(0) = 0$, with period-1 state $x$. A revision by either instrument resets mismatch to $0$; for the cleanest statics, take its total cost to be the friction $\kappa$, independent of $x$. In period 2, independently of the period-1 choice, a fresh shock arrives with probability $p$ and re-degrades the state to some $x'$ with $\theta(x') \ge \theta(x)$. With one period remaining, the agent then pays $\min\{\kappa, \theta(\cdot)\}$, revising or tolerating.

\emph{Revise now} costs $\kappa + p \min\{\kappa, \theta(x')\}$. \emph{Continue} costs $\theta(x) + p \min\{\kappa, \theta(x')\} + (1-p)\min\{\kappa, \theta(x)\}$. The new-shock branch is identical under both choices, since the shock re-degrades the state either way, so it cancels. On the candidate band $\{x : \theta(x) < \kappa\}$, continuing is optimal iff $\theta(x) + (1-p)\,\theta(x) \le \kappa$, i.e.
\[
\theta(x) \;\le\; \frac{\kappa}{2 - p},
\qquad
x_{\ell} \;=\; \theta^{-1}\!\Bigl(\frac{\kappa}{2-p}\Bigr).
\]
The band width $x_{\ell}$ is therefore strictly increasing in the revision friction $\kappa$ and in the shock probability $p$, and, replacing $\kappa$ by $\kappa + \sigma D$ for the revision's distance charge, strictly increasing in $\sigma$. The $p$-effect is the mechanism described in Section~\ref{sec:dynamic}: a likelier new shock shortens the expected life of a freshly revised plan, which lowers the return to revising now. $\qed$

\end{document}